\documentclass[11pt,letterpaper]{article}
\usepackage{emnlp2016}
\usepackage{times}
\usepackage{latexsym}

\usepackage{tabularx}
\usepackage{color}
\usepackage{verbatim}
\usepackage{graphics}
\usepackage{graphicx}
\usepackage{textcomp}

\emnlpfinalcopy

\def\emnlppaperid{***}

\title{Anchoring and Agreement in Syntactic Annotations}

 \author{Yevgeni Berzak \\ CSAIL MIT \\ berzak@mit.edu \\ 
	\And Yan Huang \\ Language Technology Lab \\ DTAL Cambridge University \\ yh358@cam.ac.uk \\
	\And Andrei Barbu \\ CSAIL MIT \\ andrei@0xab.com \\	
	\AND Anna Korhonen \\Language Technology Lab \\ DTAL Cambridge University \\ alk23@cam.ac.uk \\
	\And Boris Katz \\ CSAIL MIT \\ boris@mit.edu}

\date{}

\begin{document}
\maketitle
\begin{abstract} 
We present a study on two key characteristics of human syntactic annotations:
anchoring and agreement. Anchoring is a well known cognitive bias in 
human decision making, where judgments are drawn towards pre-existing values. 
We study the influence of anchoring on a standard approach to creation of 
syntactic resources where syntactic annotations are obtained via human 
editing of tagger and parser output. Our experiments demonstrate a clear 
anchoring effect and reveal unwanted consequences, including 
overestimation of parsing performance and lower quality of 
annotations in comparison with human-based annotations. 
Using sentences from the Penn Treebank WSJ, we also report 
systematically obtained inter-annotator agreement estimates for English 
dependency parsing. Our agreement results control for parser bias, and are 
consequential in that they are \emph{on par} with state of the art parsing 
performance for English newswire. We discuss the impact of our findings on strategies 
for future annotation efforts and parser evaluations.\footnote{The experimental 
data in this study will be made publicly available.}
\end{abstract}

\section{Introduction}
\label{sect:introduction}

Research in NLP relies heavily on the availability of human annotations
for various linguistic prediction tasks. Such resources are commonly treated 
as de facto ``gold standards'' and are used for both training and evaluation
of algorithms for automatic annotation. At the same time, human agreement on 
these annotations provides an indicator for the difficulty of the 
task, and can be instrumental for estimating upper limits for the performance 
obtainable by computational methods.

Linguistic gold standards are often constructed 
using pre-existing annotations, generated by automatic tools. 
The output of such tools is then manually corrected by human annotators to produce the
gold standard. The justification for this annotation methodology was first introduced in a set of experiments 
on POS tag annotation conducted as part of the Penn Treebank project \cite{marcus1993ptb}.
In this study, the authors concluded that tagger-based annotations 
are not only much faster to obtain, but also more consistent and of higher quality compared to 
annotations from scratch. Following the Penn Treebank, syntactic annotation projects for various languages, 
including German \cite{brants2002tiger}, French \cite{abeille2003building}, Arabic \cite{maamouri2004} 
and many others, were annotated using automatic tools as a starting point. 
Despite the widespread use of this annotation pipeline, there is, to our knowledge, little prior 
work on syntactic annotation quality and on the reliability of system evaluations on such data. 

In this work, we present a systematic study of the influence of 
automatic tool output on characteristics of annotations created for NLP purposes.
Our investigation is motivated by the hypothesis that annotations obtained using
such methodologies may be subject to the problem of \emph{anchoring}, 
a well established and robust cognitive bias in which human decisions are affected by 
pre-existing values \cite{tversky1974}. In the presence of anchors, participants reason 
relative to the existing values, and as a result may provide different solutions from those 
they would have reported otherwise. Most commonly, anchoring is 
manifested as an alignment \emph{towards} the given values. 

Focusing on the key NLP tasks of POS tagging and dependency parsing, we 
demonstrate that the standard approach of obtaining annotations via human correction 
of automatically generated POS tags and dependencies exhibits a clear anchoring effect -- a
phenomenon we refer to as \emph{parser bias}. Given this evidence, we examine 
two potential adverse implications of this effect on parser-based gold standards.

First, we show that parser bias entails substantial overestimation of parser performance. In particular, 
we demonstrate that bias towards the output of a specific tagger-parser pair 
leads to over-estimation of the performance of these tools relative to other tools.
Moreover, we observe general performance gains for automatic tools
relative to their performance on human-based gold standards. 
Second, we study whether parser bias affects the 
quality of the resulting gold standards. Extending the experimental setup 
of Marcus et al. \shortcite{marcus1993ptb}, we demonstrate that 
parser bias may lead to \emph{lower} annotation quality for 
parser-based annotations compared to human-based annotations.

Furthermore, we conduct an experiment on inter-annotator agreement for
POS tagging and dependency parsing which controls for parser bias. Our experiment 
on a subset of section 23 of the WSJ Penn Treebank yields 
agreement rates of 95.65 for POS tagging and 94.17 for dependency parsing. 
This result is significant in light of the state of the art tagging and parsing 
performance for English newswire. With parsing reaching the level of human agreement, 
and tagging surpassing it, a more thorough examination of evaluation resources 
and evaluation methodologies for these tasks is called for.  

To summarize, we present the first study to measure and analyze anchoring
in the standard parser-based approach to creation of gold standards for POS tagging
and dependency parsing in NLP. We conclude that gold standard annotations that are based on editing output
of automatic tools can lead to inaccurate figures in system evaluations and lower 
annotation quality. Our human agreement experiment, which controls for parser bias, yields
agreement rates that are comparable to state of the art automatic tagging and
dependency parsing performance, highlighting the need for a more extensive investigation
of tagger and parser evaluation in NLP. 

\section{Experimental Setup}

\subsection{Annotation Tasks}

We examine two standard annotation tasks in NLP, POS tagging and dependency parsing. 
In the POS tagging task, each word in a sentence has to be categorized 
with a Penn Treebank POS tag \cite{ptbpos} (henceforth POS). The dependency parsing 
task consists of providing a sentence with a labeled dependency tree using the
 Universal Dependencies (UD) formalism \cite{de2014universal}, according to version
1 of the UD English guidelines\footnote{http://universaldependencies.org/\#en}. To perform this 
task, the annotator is required to specify the head word index (henceforth HIND) 
and relation label (henceforth REL) of each word in the sentence.

We distinguish between three variants of these tasks, \emph{annotation}, 
\emph{reviewing} and \emph{ranking}. 
In the annotation variant, participants are asked to conduct annotation from scratch. 
In the reviewing variant, they are asked to provide alternative annotations for all
annotation tokens with which they disagree. The participants are not informed about the source of the given annotation, which, depending
on the experimental condition can be either parser output or human annotation. 
In the ranking task, the participants rank several 
annotation options with respect to their quality. Similarly to the review task, the participants
are not given the sources of the different annotation options. Participants performing the annotation, 
reviewing and ranking tasks are referred to as annotators, reviewers and judges, respectively.   

\subsection{Annotation Format}

All annotation tasks are performed using a CoNLL style text-based template,
in which each word appears in a separate line. The first two columns of each 
line contain the word index and the word, respectively. 
The next three columns are designated 
for annotation of POS, HIND and REL. 

In the annotation task, these values have to be specified by the annotator 
from scratch. In the review task, participants are required to edit pre-annotated 
values for a given sentence. The sixth 
column in the review template contains an additional \# sign, whose goal is
to prevent reviewers from overlooking and passively approving existing annotations.
Corrections are specified following this sign in a space separated format, 
where each of the existing three annotation tokens is either corrected with 
an alternative annotation value or approved using a \emph{*} sign. Approval 
of all three annotation tokens is marked by removing the \# sign. The example 
below presents a fragment from a sentence used for the reviewing task, in which
the reviewer approves the annotations of all the words,
with the exception of ``help'', where the POS is corrected from VB to NN
and the relation label \emph{xcomp} is replaced with \emph{dobj}.

\begin{footnotesize}
\texttt{
\begin{tabbing}
\hspace*{0.7cm}\=\hspace*{2.1cm}\=\hspace*{1.1cm}\=\hspace*{0.9cm}\= \kill
...	\>	\>	\>	\> \\
5	\>you	\>PRP	\>6	\>nsubj \\
6	\>need	\>VBP	\>3	\>ccomp \\
7	\>help	 \>VB	\>6	\>xcomp    \bf \# NN * dobj \\
...	\>	\>	\>	\> \\
\end{tabbing}
}
\end{footnotesize}

The format of the ranking task is exemplified below. The 
annotation options are presented to the participants in a random order. 
Participants specify the rank of each annotation token following the vertical bar. 
In this sentence, the label \emph{cop} is 
preferred over \emph{aux} for the word ``be'' and \emph{xcomp} is preferred 
over \emph{advcl} for the word ``Common''.  

\begin{footnotesize}
\texttt{
\begin{tabbing}
\hspace*{0.7cm}\=\hspace*{2cm}\=\hspace*{1.2cm}\=\hspace*{1.1cm}\= \kill
	...	\>	\>	\>	\> \\
8       \>it      \>PRP     \>10      \>nsubjpass \\
9       \>is      \>VBZ     \>10      \>auxpass \\
10      \>planed  \>VBN     \>0       \>root \\
11      \>to      \>TO      \>15      \>mark \\
12      \>be      \>VB      \>15      \> \bf aux-cop $|$ 2-1 \\
13      \>in      \>IN      \>15      \>case \\
14      \>Wimbledon       \>NNP     \>15      \>compound \\
15      \>Common	\>NNP     \>10      \> \bf advcl-xcomp $|$ 2-1 \\
	...	\>	\>	\>	\> \\
\end{tabbing}
}
\end{footnotesize}

The participants used basic validation scripts which checked for
typos and proper formatting of the annotations, reviews and rankings.
   
\subsection{Evaluation Metrics}

We measure both parsing performance and inter-annotator agreement using 
tagging and parsing evaluation metrics. This choice allows for a direct comparison 
between parsing and agreement results. In this context, POS refers to tagging
accuracy. We utilize the standard metrics Unlabeled Attachment Score (UAS)
and Label Accuracy (LA) to measure accuracy of head attachment and dependency 
labels. We also utilize the standard parsing metric Labeled Attachment Score
(LAS), which takes into account both dependency arcs and dependency labels. 
In all our parsing and agreement experiments, we exclude punctuation
tokens from the evaluation.

\subsection{Corpora}

We use sentences from two publicly available datasets, covering two different
genres. The first corpus, used in the experiments in sections \ref{bias_experiments} 
and \ref{anno_quality}, is the First Certificate in English (FCE) Cambridge Learner Corpus 
\cite{fcecorpus2011}. This dataset contains essays authored by upper-intermediate 
level English learners\footnote{The annotation bias and quality results reported 
in sections \ref{bias_experiments} and \ref{anno_quality} use the original learner 
sentences, which contain grammatical errors. These results were replicated on the 
error corrected versions of the sentences.}. 

The second corpus is the WSJ part of the Penn Treebank (WSJ PTB) 
\cite{marcus1993ptb}. Since its release, this dataset has been the most commonly 
used resource for training and evaluation of English parsers. Our experiment on 
inter-annotator agreement in section \ref{interanno} uses a random subset of the sentences 
in section 23 of the WSJ PTB, which is traditionally reserved for tagging and parsing evaluation.

\subsection{Annotators}

We recruited five students at MIT as annotators.
Three of the students are linguistics majors and two are engineering majors with linguistics minors. 
Prior to participating in this study, the annotators completed two months of 
training. During training, the students attended tutorials, and learned the annotation guidelines for PTB POS tags, 
UD guidelines, as well as guidelines for annotating challenging 
syntactic structures arising from grammatical errors. The students also 
annotated individually six practice batches of 20-30 sentences from the English Web Treebank (EWT) \cite{silveira2014gold} and FCE
corpora, and resolved annotation disagreements during group meetings.

Following the training period, the students annotated a treebank of learner 
English \cite{berzak2016} over a period of five months, three of which as a full time job. 
During this time, the students continued attending weekly meetings in which further annotation
challenges were discussed and resolved. The annotation was carried out for 
sentences from the FCE dataset, where both the original and error corrected 
versions of each sentence were annotated and reviewed. In the course of the annotation 
project, each annotator completed approximately 800 sentence annotations, and a similar 
number of sentence reviews. The annotations and reviews were done in the 
same format used in this study. With respect to our experiments, the extensive experience 
of our participants and their prior work as a group strengthen our results, as these 
characteristics reduce the effect of anchoring biases and increase inter-annotator agreement.

\section{Parser Bias}
\label{bias_experiments}

Our first experiment is designed to test whether expert human annotators are 
biased towards POS tags and dependencies generated by automatic 
tools. We examine the common out-of-domain annotation scenario, where automatic 
tools are often trained on an existing treebank in one domain, and used to generate 
initial annotations to speed-up the creation of a gold standard for a new domain. We use the 
EWT UD corpus as the existing gold standard, and a sample of the FCE dataset as the new corpus.

\subsubsection*{Procedure}

\begin{figure}[ht!]
\center
    \includegraphics[width=0.48\textwidth]{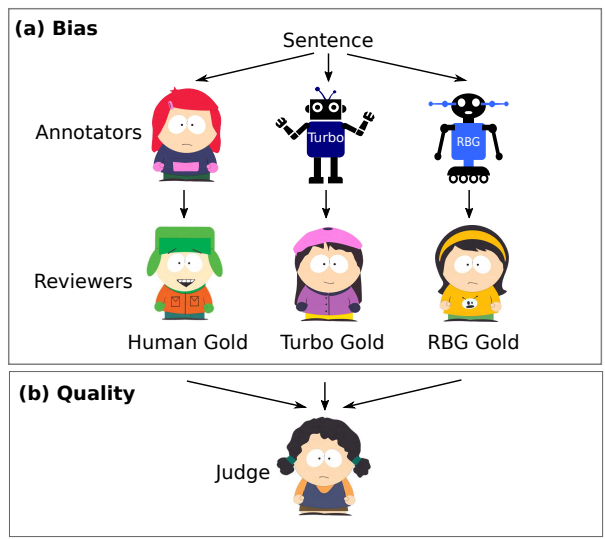}
\caption{Experimental setup for parser bias (a) and annotation quality (b) on 360 sentences (6,979 tokens) 
from the FCE. For each sentence, five human annotators are assigned at random to one of three roles: 
annotation, review or quality assessment. In the bias experiment, presented in section \ref{bias_experiments}, 
every sentence is annotated by a human, Turbo parser (based on Turbo tagger output) 
and RBG parser (based on Stanford tagger output). Each annotation is reviewed by a 
different human participant to produce three gold standards of each sentence: ``Human Gold'', 
``Turbo Gold'' and ``RBG Gold''. The fifth annotator performs a quality 
assessment task described in section \ref{anno_quality}, which requires to rank the 
three gold standards in cases of disagreement.}
\label{bias_pic}
\end{figure}

Our experimental procedure, illustrated in figure \ref{bias_pic}(a)
contains a set of 360 sentences (6,979 tokens) from the FCE, for which 
we generate three gold standards: one based on human annotations and two based on 
parser outputs. To this end, for each sentence, we assign \emph{at random} four of the 
participants to the following annotation and review tasks. The fifth participant 
is left out to perform the quality ranking task described in section \ref{anno_quality}.

The first participant annotates the sentence from scratch, and a second participant reviews 
this annotation. The overall agreement of the reviewers with the
annotators is 98.24 POS, 97.16 UAS, 96.3 LA and 94.81 LAS. 
The next two participants review parser outputs. One participant reviews 
an annotation generated by the Turbo tagger and parser \cite{martins2013}. 
The other participant reviews the output of the Stanford tagger 
\cite{toutanova2003tagger} and RBG parser \cite{lei2014}. The 
taggers and parsers were trained on the gold annotations of the EWT UD 
treebank, version 1.1. Both parsers use predicted POS tags for the FCE sentences. 

\begin{table*}[ht!]
\begin{center}
\small
\begin{tabular}{lllll|llll}
            & \multicolumn{4}{c}{\bf \underline{Turbo}} & \multicolumn{4}{c}{\bf \underline{RBG}}\\  
            & POS & UAS & LA & LAS & POS & UAS & LA & LAS \\ \cline{2-9}
 \bf Human Gold & 95.32 & 87.29	& 88.35	& 82.29	& 95.59	& 87.19	& 88.03	& 82.05\\ \hline
 \bf Turbo Gold & \bf 97.62 & \bf 91.86 & \bf 92.54 & \bf 89.16 & \it 96.64 & \it 89.16 & \it 89.75 & \it 84.86 \\ 
 Error Reduction \%   &  49.15 & 35.96 & 35.97 & 38.79 & 23.81 & 15.38 & 14.37 & 15.65 \\ \hline	
 \bf RBG Gold & 96.43 & \it 88.65& \it 89.95& \it 84.42	& \bf 97.76& \bf 91.22& \bf 91.84& \bf 87.87 \\ 
 Error Reduction \% & 23.72	&10.7&	13.73&	12.03&	49.21&	31.46&	31.83&	32.42 \\
\end{tabular}
\end{center}
\caption{Annotator bias towards taggers and parsers on 360 sentences (6,979 tokens) 
from the FCE. Tagging and parsing results are reported for the Turbo parser (based on the output 
of the turbo Tagger) and RBG parser (based on the output of the Stanford tagger) on three gold standards. 
Human Gold are manual corrections of 
human annotations. Turbo Gold are manual corrections of the output
of Turbo tagger and Turbo parser. RBG Gold are manual corrections of 
the Stanford tagger and RBG parser. Error reduction rates are reported 
relative to the results obtained by the two tagger-parser pairs on the 
Human Gold annotations. 
Note that (1) The parsers perform equally well on Human Gold. (2) Each parser performs 
better than the other parser on its own reviews. 
(3) Each parser performs better on the reviews of the other parser compared to its performance
on Human Gold. The differences in (2) and (3) are statistically significant with $p\ll0.001$ using
McNemar's test.
}\label{parser_bias_table}
\end{table*}

Assigning the reviews to the human annotations yields a human 
based gold standard for each sentence called ``Human Gold''. Assigning the 
reviews to the tagger and parser outputs yields two parser-based 
gold standards, ``Turbo Gold'' and ``RBG Gold''. 
We chose the Turbo-Turbo and Stanford-RBG tagger-parser pairs as 
these tools obtain comparable 
performance on standard evaluation benchmarks, while yielding substantially 
different annotations due to different training algorithms and feature sets. 
For our sentences, the agreement between the Turbo tagger and 
Stanford tagger is 96.97 POS. The agreement between the 
Turbo parser and RBG parser based on the respective tagger 
outputs is 90.76 UAS, 91.6 LA and 87.34 LAS.

\subsubsection*{Parser Specific and Parser Shared Bias} 
In order to test for parser bias, in table \ref{parser_bias_table} 
we compare the performance of the Turbo-Turbo and Stanford-RBG 
tagger-parser pairs on our three gold standards.
First, we observe that while these tools perform equally 
well on Human Gold, each tagger-parser pair 
performs better than the other on its own reviews.
These \emph{parser specific} performance gaps are substantial, with an 
average of 1.15 POS, 2.63 UAS, 2.34 LA and 3.88 LAS between the two conditions.
This result suggests the presence of a bias towards the output of 
specific tagger-parser combinations. The practical implication of 
this outcome is that a gold standard created by editing an output of a parser is
likely to boost the performance of that parser in evaluations and
over-estimate its performance relative to other parsers.

Second, we note that the performance of each of the parsers on 
the gold standard of the other parser is still higher than its performance 
on the human gold standard. The average performance gap between these 
conditions is 1.08 POS, 1.66 UAS, 1.66 LA and 2.47 LAS.   
This difference suggests an annotation bias towards \emph{shared} aspects 
in the predictions of taggers and parsers, which differ from the human 
based annotations. The consequence of this observation is that irrespective 
of the specific tool that was used to pre-annotate the data, parser-based 
gold standards are likely to result in higher parsing performance relative 
to human-based gold standards.

Taken together, the parser specific and parser shared effects lead to 
a dramatic overall average error reduction of 49.18\% POS, 33.71\% UAS, 
34.9\% LA and 35.61\% LAS on the parser-based gold standards compared to the 
human-based gold standard. To the best of our knowledge, these
results are the first systematic demonstration of the tendency of the common 
approach of parser-based creation of gold standards to yield biased annotations and 
lead to overestimation of tagging and parsing performance.

\section{Annotation Quality}
\label{anno_quality}

In this section we extend our investigation
to examine the impact of parser bias on the quality of parser-based gold 
standards. To this end, we perform a manual comparison between 
human-based and parser-based gold standards.

Our quality assessment experiment, depicted schematically in figure 
\ref{bias_pic}(b), is a ranking task. For each sentence, a randomly chosen judge, 
who did not annotate or review the given sentence, ranks disagreements 
between the three gold standards Human Gold, 
Turbo Gold and RBG Gold, generated in the parser bias experiment in 
section \ref{bias_experiments}. 

\begin{table}[ht!]
\resizebox{\columnwidth}{!}{%
\begin{tabular}{lllll}
Human Gold Preference \% & POS & HIND & REL \\ \hline
Turbo Gold & 64.32* & 63.96* & 61.5* \\
\# disagreements & 199 & 444 & 439 \\ \hline
RBG Gold  & 56.72 & 61.38* & 57.73*  \\ 
\# disagreements & 201 & 435  & 440 \\ 
\end{tabular}%
}
\caption{Human preference rates for a human-based gold standard Human Gold over the two 
parser-based gold standards Turbo Gold and RBG Gold.
\# disagreements denotes the number of tokens that differ between 
Human Gold and the respective parser-based gold standard.
Statistically significant values for a two-tailed Z test with $p<0.01$ are marked with *.
Note that for both tagger-parser pairs, human judges tend to prefer human-based over parser-based annotations.
}
\label{quality_table}
\end{table}

Table \ref{quality_table} presents the preference rates of judges for the 
human-based gold standard over each of the two parser-based gold standards.
In all three evaluation categories, human judges tend to prefer the human-based 
gold standard over both parser-based gold standards. This result demonstrates
that the initial reduced quality of the parser outputs compared to human annotations 
indeed percolates via anchoring to the resulting gold standards. 

The analysis of the quality assessment experiment thus far did not 
distinguish between cases where the two parsers agree and where 
they disagree. In order to gain further insight into 
the relation between parser bias and annotation quality, we break
down the results reported in table \ref{quality_table} into two cases 
which relate directly to the \emph{parser specific} and \emph{parser shared} 
components of the tagging and parsing performance gaps observed in 
the parser bias results reported in section \ref{bias_experiments}. 
In the first case, called ``parser specific approval'', a reviewer
approves a parser annotation which disagrees both with the output of
the other parser and the Human Gold annotation. In the second case, 
called ``parser shared approval'', a reviewer approves a parser output 
which is shared by both parsers but differs with respect to Human Gold. 

\begin{table}[ht!]
\resizebox{\columnwidth}{!}{%
\begin{tabular}{lllll}
Human Gold Preference \% & POS & HIND & REL \\ \hline
Turbo specific approval  & 85.42* & 78.69* & 80.73*  \\ 
\# disagreements & 48 & 122  & 109 \\ \hline
RBG specific approval & 73.81* & 77.98* & 77.78*  \\ 
\# disagreements & 42 & 109  & 108 \\ \hline
Parser shared approval & 51.85 & 58.49* & 51.57  \\ 
\# disagreements & 243 & 424  & 415 \\ 

\end{tabular}%
}
\caption{Breakdown of the Human preference rates for the human-based gold 
standard over the parser-based gold standards in table \ref{quality_table}, 
into cases of agreement and disagreement between the two parsers. 
Parser specific approval are cases in which a parser output 
approved by the reviewer differs from both the output of the other 
parser and the Human Gold annotation. Parser shared approval denotes cases 
where an approved parser output is identical to the output of the other 
parser but differs from the Human Gold annotation. Statistically significant values
for a two-tailed Z test with $p<0.01$ are marked with *. Note that parser specific 
approval is substantially more detrimental to the resulting annotation quality 
compared to parser shared approval.
}
\label{specific_shared_rankings}
\end{table}

Table \ref{specific_shared_rankings} presents the judge preference rates 
for the Human-Gold annotations in these two scenarios. We observe 
that cases in which the parsers disagree are of substantially worse 
quality compared to human-based annotations. 
However, in cases of agreement 
between the parsers, the resulting gold standards do not exhibit 
a clear disadvantage relative to the Human Gold annotations. 

This result highlights the crucial role of parser specific approval 
in the overall preference of judges towards human-based annotations 
in table \ref{quality_table}. Furthermore, it suggests that annotations 
on which multiple state of the art parsers agree are of sufficiently high 
accuracy to be used to save annotation time without substantial impact 
on the quality of the resulting resource. In section \ref{discussion} we 
propose an annotation scheme which leverages this insight.

\section{Inter-annotator Agreement}
\label{interanno}

Agreement estimates in NLP are often obtained in annotation
setups where both annotators edit the same automatically generated input. 
However, in such experimental conditions, anchoring can
introduce cases of spurious disagreement as well as spurious 
agreement between annotators due to alignment of one or both participants 
towards the given input. The initial quality of the provided annotations
in combination with the parser bias effect observed in section \ref{bias_experiments} 
may influence the resulting agreement estimates. 
For example, in Marcus et al. \shortcite{marcus1993ptb} annotators were 
shown to produce POS tagging agreement of 92.8 on annotation from scratch, 
compared to 96.5 on reviews of tagger output.

Our goal in this section is to obtain estimates for inter-annotator agreement 
on POS tagging and dependency parsing that control for parser bias, 
and as a result, reflect more accurately human 
agreement on these tasks. We thus introduce a novel pipeline based on human annotation only, 
which eliminates parser bias from the agreement measurements.
Our experiment extends the human-based annotation study of Marcus et al. \shortcite{marcus1993ptb} to include 
also syntactic trees. Importantly, we include an additional review step for the initial annotations, 
designed to increase the precision of the agreement
measurements by reducing the number of errors in the original annotations.

\begin{figure}[ht!]  
\begin{center}
    \includegraphics[width=0.25\textwidth]{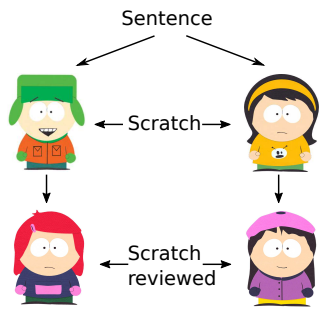}
\caption{Experimental setup for the inter-annotator agreement experiment.
	300 sentences (7,227 tokens) from section 23 of the PTB-WSJ are annotated and reviewed by four participants.
	The participants are assigned to the following tasks \emph{at random} for each sentence.	
	Two participants annotate the sentence from scratch, and the remaining two participants
	review one of these annotations each.
	Agreement is measured on the annotations (``scratch'') as well after assigning the review edits (``scratch reviewed'').}
\end{center}
\label{agr_pic}
\end{figure}

For this experiment, we use 300 sentences (7,227 tokens) 
from section 23 of the PTB-WSJ, the standard test set for English 
parsing in NLP. The experimental setup, depicted graphically in figure 2, 
includes four participants randomly assigned for each sentence to annotation and review tasks. Two of the 
participants provide the sentence with annotations from scratch, while the 
remaining two participants provide reviews. Each reviewer edits one of the annotations
independently, allowing for correction of annotation errors while maintaining the independence
of the annotation sources. 
We measure agreement between the initial annotations (``scratch''), as 
well as the agreement between the reviewed versions of our sentences (``scratch reviewed'').

The agreement results for the annotations and the reviews
are presented in table \ref{agreement_table}. The initial agreement rate on 
POS annotation from scratch is higher than in \cite{marcus1993ptb}. 
This difference is likely to arise, at least in part, due to the fact that 
their experiment was conducted at the beginning of the annotation project,
when the annotators had a more limited annotation experience compared to our 
participants. Overall, we note that the agreement rates from scratch are relatively 
low. The review round raises the agreement on all the evaluation categories due 
to elimination of annotation errors present the original annotations. 

\begin{table}[ht!] 
\center
\small
\begin{tabular}{lllll}
             & POS & UAS & LA & LAS \\ \hline
  scratch  & 94.78  & 93.07 &  92.3 &  88.32 \\
  scratch reviewed  & \bf 95.65 & \bf 94.17 & \bf 94.04 & \bf  90.33 \\ 
\end{tabular}
\caption{Inter-annotator agreement on 300 sentences (7,227 tokens) from the PTB-WSJ section 23.
``scratch'' is agreement on independent annotations from scratch. ``scratch reviewed'' is agreement 
 on the same sentences after an additional independent review round of the annotations.}
\label{agreement_table}
\end{table}

Our post-review agreement results are consequential in light of the current state 
of the art performance on tagging and parsing in NLP. For more than a decade, POS 
taggers have been achieving over 97\% accuracy with the PTB POS tag set on the PTB-WSJ
test set. For example, the best model of the Stanford tagger reported in Toutanova et al. \shortcite{toutanova2003tagger} 
produces an accuracy of 97.24 POS on sections 22-24 of the PTB-WSJ. These accuracies 
are above the human agreement in our experiment.

With respect to dependency parsing, recent parsers obtain results which
are on par or higher than our inter-annotator agreement estimates. For example, Weiss et al. \shortcite{weiss-etAl:2015:ACL}
report 94.26 UAS and Andor et al. \shortcite{andor2016} report 94.61 UAS on section 23 of the PTB-WSJ using an automatic 
conversion of the PTB phrase structure trees to Stanford dependencies \cite{de2006generating}.
These results are not fully comparable to ours due to differences in the utilized dependency formalism and the
automatic conversion of the annotations. 
Nonetheless, we believe that the similarities in the tasks and evaluation data are 
sufficiently strong to indicate that dependency parsing for standard English newswire
may be reaching human agreement levels. 

\section{Related Work}

The term ``anchoring'' was coined in a seminal paper by Tversky and Kahneman 
\shortcite{tversky1974}, which demonstrated that numerical estimation can be biased
by uninformative prior information. Subsequent work 
across various domains of decision making confirmed the robustness of anchoring 
using both informative and uninformative anchors  \cite{furnham2011literature}. 
Pertinent to our study, anchoring biases were also demonstrated 
when the participants were domain experts, although to a lesser degree 
than in the early anchoring experiments \cite{wilson1996,mussweiler2000numeric}. 

Prior work in NLP examined the influence of pre-tagging \cite{fort2010} and 
pre-parsing \cite{skjaerholt2013} on human annotations.
Our work introduces a systematic study of this topic using a novel experimental 
framework as well as substantially more sentences and annotators. 
Differently from these studies, our methodology enables characterizing annotation bias 
as anchoring and measuring its effect on tagger and parser evaluations. 

Our study also extends the POS tagging experiments of Marcus et al. 
\shortcite{marcus1993ptb}, which compared inter-annotator agreement and 
annotation quality on manual POS tagging in annotation from scratch and 
tagger-based review conditions. The first result reported in that study 
was that tagger-based editing increases inter-annotator agreement compared 
to annotation from scratch. Our work provides a novel agreement benchmark 
for POS tagging which reduces annotation errors through a review process while controlling 
for tagger bias, and obtains agreement measurements for dependency parsing. 
The second result reported in Marcus et al. \shortcite{marcus1993ptb} was that tagger-based 
edits are of higher quality compared to annotations from scratch when evaluated against 
an additional independent annotation. We modify this experiment by introducing ranking as an alternative mechanism 
for quality assessment, and adding a review round for human annotations from scratch. 
Our experiment demonstrates that in this configuration, parser-based annotations 
are of \emph{lower} quality compared to human-based annotations.

Several estimates of expert inter-annotator agreement for English 
parsing were previously reported. However, most such evaluations were conducted using annotation 
setups that can be affected by an anchoring bias \cite{carroll1999corpus,rambow2002dependency,silveira2014gold}.
A notable exception is the study of Sampson and Babarczy \shortcite{sampson2008definitional} who measure agreement on annotation
from scratch for English parsing in the SUSANNE framework \cite{sampson1995english}. 
The reported results, however, are not directly comparable to ours, due to the use of a 
substantially different syntactic representation, as well as a different agreement metric. Their study further 
suggests that despite the high expertise of the annotators, the main source of annotation disagreements was 
annotation errors. Our work alleviates this issue by using annotation reviews, which reduce 
the number of erroneous annotations while maintaining the independence of the annotation sources.
Experiments on non-expert dependency annotation from scratch were previously reported
for French, suggesting low agreement rates (79\%) with an expert annotation benchmark \cite{gerdes2013}.

\section{Discussion}
\label{discussion}

We present a systematic study of the impact of anchoring on POS and dependency
annotations used in NLP, demonstrating that annotators exhibit an anchoring bias 
effect towards the output of automatic annotation tools. 
This bias leads to an artificial boost of performance figures for the parsers in question
and results in lower annotation quality as compared with human-based annotations.

Our analysis demonstrates that despite the adverse effects of parser bias, 
predictions that are shared across different parsers do not significantly lower the 
quality of the annotations. This finding gives rise to the following 
\emph{hybrid annotation} strategy as a potential future alternative to human-based 
as well as parser-based annotation pipelines.
In a hybrid annotation setup, human annotators review annotations on which several 
parsers agree, and complete the remaining annotations from scratch. Such a strategy 
would largely maintain the annotation speed-ups of parser-based annotation schemes.
At the same time, it is expected to achieve annotation quality comparable to human-based 
annotation by avoiding parser specific bias, which plays a pivotal role in the reduced 
quality of single-parser reviewing pipelines.  

Further on, we obtain, to the best of our knowledge for the first time, syntactic inter-annotator 
agreement measurements on WSJ-PTB sentences. Our experimental procedure reduces annotation 
errors and controls for parser bias. Despite the detailed annotation guidelines, the 
extensive experience of our annotators, and their prior work as a group, our experiment 
indicates rather low agreement rates, which are below state of the art tagging performance 
and on par with state of the art parsing results on this dataset. 
We note that our results do not necessarily reflect an upper bound on the achievable 
syntactic inter-annotator agreement for English newswire. Higher agreement rates 
could in principle be obtained through further annotator training, refinement and revision 
of annotation guidelines, as well as additional automatic validation tests for the annotations. 
Nonetheless, we believe that our estimates reliably reflect a realistic scenario
of expert syntactic annotation.

The obtained agreement rates call for a more extensive examination of annotator 
disagreements on parsing and tagging. Recent work in this area has already proposed an analysis
of expert annotator disagreements for POS tagging in the absence of annotation guidelines \cite{plank2014}. 
Our annotations will enable conducting such studies for annotation with guidelines, 
and support extending this line of investigation to annotations of syntactic dependencies.
As a first step towards this goal, we plan to carry out an in-depth analysis 
of disagreement in the collected data, characterize 
the main sources of inconsistent annotation and subsequently formulate 
further strategies for improving annotation accuracy. We believe that better understanding of 
human disagreements and their relation to disagreements between humans and parsers will 
also contribute to advancing evaluation methodologies for POS tagging and syntactic 
parsing in NLP, an important topic that has received only limited attention thus 
far \cite{schwartz2011,plank2015}.

Finally, since the release of the Penn Treebank in 1992, it has been serving as the standard
benchmark for English parsing evaluation. Over the past few years, improvements in parsing
performance on this dataset were obtained in small increments, and are commonly reported
without a linguistic analysis of the improved predictions. As dependency parsing performance 
on English newswire may be reaching human expert agreement, not only new evaluation practices,
but also more attention to noisier domains and other languages 
may be in place.

\section*{Acknowledgments}
We thank our terrific annotators Sebastian Garza, Jessica Kenney, Lucia Lam, Keiko Sophie Mori 
and Jing Xian Wang. We are also grateful to Karthik Narasimhan and the anonymous reviewers for valuable 
feedback on this work. This material is based upon work supported by the 
Center for Brains, Minds, and Machines (CBMM) funded by NSF STC award CCF-1231216. This work
was also supported by AFRL contract No. FA8750-15-C-0010 and by ERC Consolidator Grant LEXICAL (648909).


\bibliographystyle{emnlp2016}
\bibliography{emnlp2016}

\end{document}